%% file: main.tex
\documentclass[10pt,twocolumn,letterpaper]{article}

\usepackage{iccv}              


\usepackage{threeparttable}
\usepackage{colortbl}
\usepackage{adjustbox}
\definecolor{iccvblue}{rgb}{0.21,0.49,0.74}
\usepackage[pagebackref,breaklinks,colorlinks,allcolors=iccvblue]{hyperref}

\def\paperID{4912} 
\def\confName{ICCV}
\def\confYear{2025}

\title{	
Bridging Domain Generalization to Multimodal Domain Generalization via Unified Representations}

\author{Hai Huang$^{1}$ \quad Yan Xia$^{1}$ \quad Sashuai Zhou$^{1}$\quad Hanting Wang$^{1}$\quad Shulei Wang$^{1}$ \quad Zhou Zhao$^{1\dagger}$
\\
$^{1}$ Zhejiang University\\
{\tt\small haihuangcode@outlook.com}\quad
{\tt\small zhaozhou@zju.edu.cn}
}

\usepackage[misc]{ifsym}
\newcommand\blfootnote[1]{
    \begingroup
    \renewcommand\thefootnote{}\footnote{#1}
    \addtocounter{footnote}{-1}
    \endgroup
}
\usepackage[hang]{footmisc}

\begin{document}
\maketitle

{
    \blfootnote{$^\dagger$Corresponding author}
}

\input{sec/0_abstract}    
\input{sec/1_intro}
\input{sec/2_relatedwork}
\input{sec/3_method}
\input{sec/4_experiment}
\input{sec/5_conclusion}
{
    \small
    \bibliographystyle{ieeenat_fullname}
    \bibliography{main}
}
\newpage
\input{sec/X_supplyment}

\end{document}

%% file: sec/0_abstract.tex
\begin{abstract}
Domain Generalization (DG) aims to enhance model robustness in unseen or distributionally shifted target domains through training exclusively on source domains. Although existing DG techniques, such as data manipulation, learning strategies, and representation learning, have shown significant progress, they predominantly address single-modal data. With the emergence of numerous multi-modal datasets and increasing demand for multi-modal tasks, a key challenge in Multi-modal Domain Generalization (MMDG) has emerged: enabling models trained on multi-modal sources to generalize to unseen target distributions within the same modality set.
Due to the inherent differences between modalities, directly transferring methods from single-modal DG to MMDG typically yields sub-optimal results. These methods often exhibit randomness during generalization due to the invisibility of target domains and fail to consider inter-modal consistency. Applying these methods independently to each modality in the MMDG setting before combining them can lead to divergent generalization directions across different modalities, resulting in degraded generalization capabilities. To address these challenges, we propose a novel approach that leverages Unified Representations to map different paired modalities together, effectively adapting DG methods to MMDG by enabling synchronized multi-modal improvements within the unified space. Additionally, we introduce a supervised disentanglement framework that separates modal-general and modal-specific information, further enhancing the alignment of unified representations. Extensive experiments on benchmark datasets, including EPIC-Kitchens and Human-Animal-Cartoon, demonstrate the effectiveness and superiority of our method in enhancing multi-modal domain generalization.
\end{abstract}

%% file: sec/1_intro.tex
\vspace{-5mm}
\section{Introduction}
\label{sec:intro}


Domain Generalization (DG) aims to enable models trained on multiple source domains to maintain robust performance in unseen or distributionally different target domains, thereby enhancing their adaptability and practicality in real-world scenarios. Addressing domain shift has become a crucial research topic in various fields, including healthcare~\cite{li2020domain, liu2021feddg, ouyang2022causality}, autonomous driving~\cite{sanchez2023domain, choi2021robustnet}, and person re-identification~\cite{bai2021person30k}. Existing DG methods can be broadly categorized into three main approaches: \textit{data manipulation}~\cite{zhang2017mixup, tobin2017domain, zhou2020deep}, \textit{learning strategies}~\cite{carlucci2019domain, huang2020self, li2018learning}, and \textit{representation learning}~\cite{pan2018two, tzeng2014deep, ganin2016domain}. These approaches have made significant progress in recent years. However, most studies primarily focus on single-modal data. With the emergence of multimodal datasets and the growing demand for multimodal tasks, Multi-modal Domain Generalization (MMDG) has become an emerging challenge. MMDG aims to fully exploit and leverage the complementary information across different modalities to enhance the generalization capability of each modality in unseen domains.

\begin{figure}
    \centering
    \includegraphics[width=1.0\linewidth]{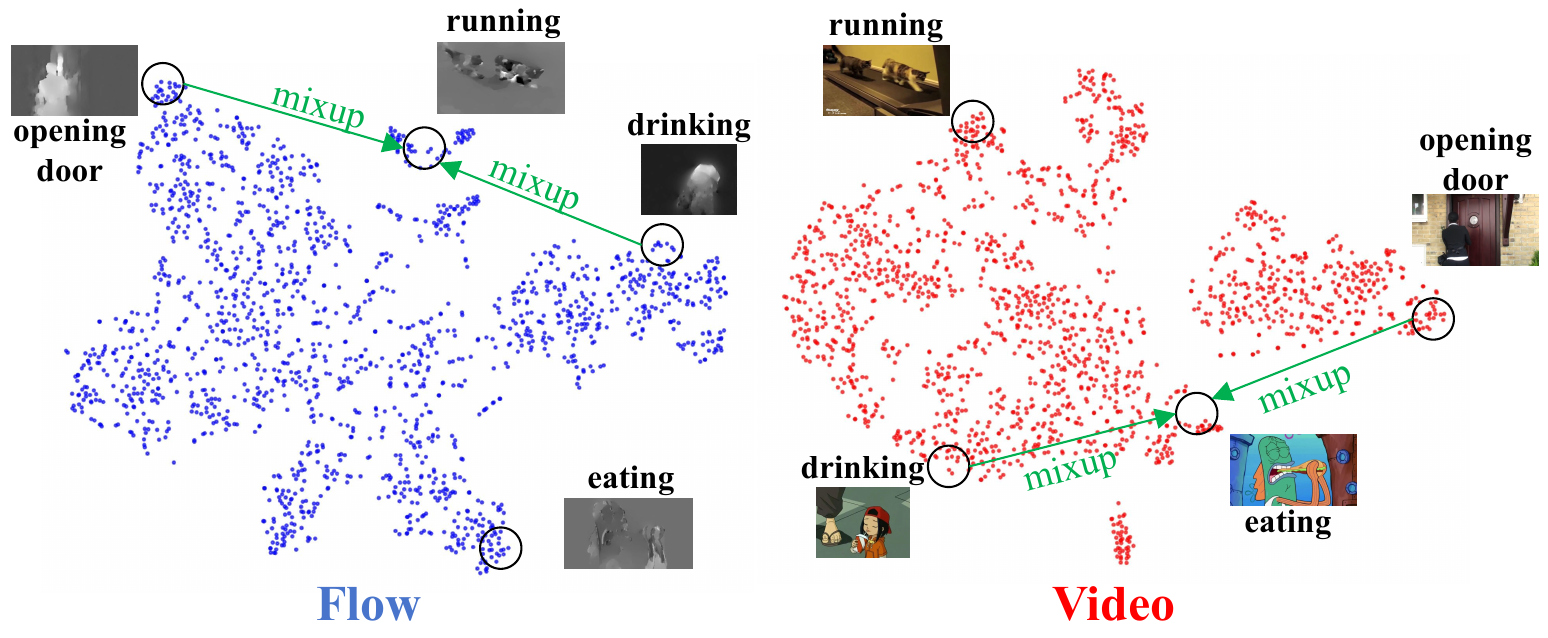}
    \caption{Illustration of conflicts when applying Mixup across different modalities.}
    \label{fig:video_flow}
    \vspace{-5mm}
\end{figure}

Among various data augmentation methods, Mixup~\cite{zhang2017mixup} is widely used for its simplicity and effectiveness in enhancing model generalization. By interpolating inputs and labels of two random samples, it smooths decision boundaries and reduces overfitting, making it particularly effective for DG tasks. However, directly applying Mixup to MMDG often underperforms methods that consider multimodal interactions~\cite{dong2023simmmdg,CMRF}. Although paired multimodal data share the same category set (e.g., “cat”, “dog”, “airplane” in image, text, and video), variations in distribution, noise, and context dependencies can cause semantic conflicts or feature misalignment when Mixup is applied uniformly across modalities. Figure~\ref{fig:video_flow} illustrates that Mixup on the same two categories in Flow and Video modalities can produce an interpolated result semantically closer to an different third category (the interpolated result of Flow leans toward “running”, while that of Video leans toward “eating”). Other common DG methods, such as normalization techniques~\cite{pan2018two} and self-supervised learning~\cite{carlucci2019domain}, also face limitations due to multimodal discrepancies.

As shown in Table~\ref{tab:modal_compete}, directly applying methods such as JiGen~\cite{carlucci2019domain} and Mixup~\cite{zhang2017mixup} improves both DG and MMDG performance. However, compared to the improvements these methods achieve in single-modal settings, their enhancement in multi-modal joint training is significantly smaller. For instance, JiGen (V) improves from 58.73\% to 61.60\% over Base (V), whereas JiGen (VAF) only increases by 1.61\% compared to Base (VAF). These results indicate that the intrinsic properties of multi-modal data constrain the direct transferability of DG methods to MMDG, often leading to suboptimal performance. Moreover, it is evident that for Base, JiGen, and Mixup, the single-modal performance in the multi-modal joint training setting is consistently inferior to that of models trained exclusively on a single modality. This indicates the presence of modality asynchrony~\cite{huang2022modality}, where competition among different modalities restricts the effective learning of individual modality features.

To address this, we propose mapping semantically aligned features from different modalities into a unified representation space~\cite{radford2021learning, girdhar2023imagebind,xia2024achieving}, where Mixup, self-supervised learning, and other DG techniques can be applied. This enables synchronized multimodal DG, effectively improving generalization in the MMDG setting.

\input{table/modal_compete}

A common approach for constructing a unified representation is contrastive learning~\cite{radford2021learning, girdhar2023imagebind}. However, directly applying contrastive learning in MMDG introduces new challenges, as domain shift must also be considered. Even when modalities share the same categories, variations in data collection methods, background conditions, and styles across domains further amplify distributional discrepancies. For instance, images of cats taken indoors versus outdoors, or audio recorded using different cameras or sensors, belong to the same category (e.g., "cat" or a specific sound) yet exhibit significant feature differences. Applying contrastive learning indiscriminately across all domains and modalities may introduce excessive noise, weakening generalization. Conversely, performing contrastive learning separately for each modality within each domain leads to computational overhead and fragmented semantics, resulting in a dispersed representation space lacking semantic coherence.

To address this, we design a disentanglement framework based on supervised contrastive learning~\cite{khosla2020supervised} that aligns information solely related to category semantics while separating it from domain-specific and modality-specific information. Our framework disentangles any input, regardless of domain or modality, into two components: general information, which captures category-related semantics, and specific information, which contains domain-specific and modality-specific features. The general information of samples from the same category is encouraged to be close, while each sample's general and specific information remain distinct. We then apply Mixup~\cite{zhang2017mixup}, JiGen~\cite{carlucci2019domain} and IBN-Net~\cite{pan2018two} in a unified representation space that contains only general information, ensuring synchronized enhancement across modalities. The enriched unified representation is subsequently combined with the previously separated specific information for model training, significantly improving generalization. Essentially, this approach reformulates the MMDG problem into a DG problem, making it more tractable. Our contributions are as follows:  
\begin{itemize}
\item We analyze why directly applying DG methods to MMDG leads to suboptimal performance and provide empirical validation.  
\item We propose leveraging a unified representation as a bridge between DG and MMDG and design a disentanglement framework based on supervised contrastive learning and mutual information decoupling to construct an effective multimodal unified representation.  
\item Our method achieves state-of-the-art performance on MMDG benchmarks such as EPIC and HAC, effectively extending previously successful DG approaches to the MMDG setting.
\end{itemize}

%% file: table/modal_compete.tex
\begin{table}[]
    \centering
    \setlength{\tabcolsep}{10pt}
    \resizebox{0.9\linewidth}{!}{
    \begin{tabular}{lcccc}
\toprule
& \multicolumn{4}{c}{\textbf{EPIC-Kitchens}}  \\
\cmidrule(lr){2-5} \textbf{Method} & Video & Audio & Flow & V-A-F\\
\midrule
Base (V) &  58.73 & - & - & -\\
Base (A) &  - & 40.04 & - & -\\
Base (F) &  - & - & 57.61 & -\\
Base (VAF) &  57.13 & 37.96 & 56.65 & 61.20\\
\midrule
JiGen (V) & \underline{61.60} & - & - & -\\
JiGen (A) &  - &  \underline{42.06} & - & -\\
JiGen (F) &  - & - &  \underline{59.85} & -\\
JiGen (VAF) &  59.23 & 39.58 & 57.18 & \underline{62.81}\\
\midrule
Mixup (V) &  \textbf{61.92} & - & - & -\\
Mixup (A) &  - & \textbf{42.74} & - & -\\
Mixup (F) &  - & - & \textbf{60.16} & -\\
Mixup (VAF) &  58.98 & 40.67 & 58.25 & \textbf{63.42}\\
\bottomrule
    \end{tabular}}
    \caption{MMDG analysis on EPIC-Kitchens using video, audio, and flow data, with results reported in accuracy (\%). 'Base' denotes vanilla multi-modal joint training without domain generalization. Parentheses indicate training modalities: V (Video), A (Audio), and F (Flow). 'Video,' 'Audio,' 'Flow,' and 'V-A-F' represent uni-modal and multi-modal testing. Results are averaged by using each domain as target.
}
    \label{tab:modal_compete}
\vspace{-5mm}
\end{table}

%% file: sec/2_relatedwork.tex
\section{Related work}
\label{sec:relatedwork}
\subsection{Domain Generalization} 
Domain Generalization (DG) aims to train models on multiple source domains to ensure effective generalization to unseen target domains, with the constraint that target domain data is unavailable during training. Previous research has identified three primary categories of DG methods~\cite{wang2022generalizing}: data manipulation, learning strategies, and representation learning. Data manipulation techniques enhance generalization by increasing the diversity of the training data. Classic works such as Mixup~\cite{zhang2017mixup} generate new instances by linearly interpolating between data and labels, while other approaches use simulated environments to generate additional data~\cite{tobin2017domain} or employ adversarial training to synthesize data from previously unseen domains~\cite{zhou2020deep}. Learning strategies focus on self-supervised learning~\cite{carlucci2019domain}, gradient-based operations~\cite{huang2020self}, and meta-learning~\cite{li2018learning}, all of which aim to improve the model's ability to adapt to new domains. On the other hand, representation learning methods focus on learning domain-invariant representations through techniques such as instance normalization ~\cite{pan2018two}, explicit feature distribution alignment~\cite{tzeng2014deep,wang2025irbridge}, and domain-adversarial neural networks~\cite{ganin2016domain}. These methods are designed to improve the model's capacity to generalize across diverse domains.

\subsection{Multi-modal Domain Generalization} 
In the field of multimodal learning, research on domain generalization (DG) has made preliminary progress. SimMMDG~\cite{dong2023simmmdg} enhances model generalization through generation, leveraging contrastive learning but without explicitly designing for a unified representation. It primarily utilizes contrastive learning to improve multimodal representation capacity. RNA-Net~\cite{planamente2022domain} introduces a relative norm alignment loss to balance the norms of audio and video features, addressing the multimodal domain generalization challenge. CMRF~\cite{CMRF} integrates Sharpness-Aware Minimization~\cite{foret2020sharpness} with multimodal learning to mitigate modality competition, highlighting the impact of modality competition on transferring single-modal DG methods to MMDG. However, it only explores a specific approach rather than proposing a generalized paradigm. To the best of our knowledge, no prior work has explored multimodal domain generalization via unified representations, making our approach the first attempt in this direction.

\subsection{Multi-modal Unified Representation} 
Multi-modal unified representation has recently emerged as a promising area of research, aiming to align disparate modalities into a shared latent space. This approach can be broadly categorized into two types: implicit unified representation \cite{petridis2018audio, andonian2022robust, chen2020uniter, wang2025towards, huang2025overcoming} and explicit unified representation \cite{zhao2022towards, lu2022unified, liu2021cross, xia2024achieving, fang2024ace, hdcid, huang2025semantic}. The former focuses on leveraging contrastive learning to bring different modalities closer in the latent space \cite{sarkar2024xkd, andonian2022robust}, or employs a general modality-agnostic encoder to encode multiple modalities \cite{wang2022vlmixer}, while the latter utilizes techniques such as optimal transport \cite{duan2022multi}, vector quantization \cite{liu2021cross}, and other methods to map information from various modalities onto a set of universal prototypes or dictionary vectors. These approaches have been applied to a range of modality combinations, including image-text \cite{duan2022multi}, audio-video \cite{liu2021cross}, and speech-text \cite{han2021learning}. In contrast to these efforts, our work is the first to apply Multi-modal Unified Representation to address MMDG, and we propose a supervised decoupling framework that is particularly well-suited for MMDG tasks, offering a more effective solution than previous unified representation methods.

%% file: sec/3_method.tex
\section{Method}
\label{sec:method}



\subsection{Preliminaries}
We selected one commonly used method from each of the three categories of DG methods. For Data Manipulation, we chose Mixup~\cite{zhang2017mixup}; for Learning Strategy, we adopted JiGen~\cite{carlucci2019domain}, a well-established self-supervised learning method; and for Representation Learning, we used IBN-Net~\cite{pan2018two}, a regularization approach. Both Mixup and JiGen were originally proposed for raw image inputs (single modality), but subsequent multimodal improvements \cite{hao2023mixgen,dong2024towards} have focused on operating at the representation level, achieving remarkable results. We also conducted experiments on both raw inputs and representations in the multimodal setting, and found that representations outperformed raw inputs. Consequently, all subsequent baselines were implemented based on representations.

\noindent
\textbf{Mixup: }
Given a set of paired multi-modal data \( \mathbb{X} = \{(\mathbf{x}_{i}^{A}, \mathbf{x}_{i}^{B}, \mathbf{x}_{i}^{C})\}_{i=1}^{N_{1}} \), with size \( N \), where \( A \), \( B \), and \( C \) represent different modalities, each modality may contain multiple domains, i.e., \( \mathbf{x}^{A} = \{\mathbf{x}^{A,D1}, \mathbf{x}^{A,D2}, \dots\} \). In multi-modal mixup, the pairing samples across modalities are mixed consistently with the same \( \lambda \). Specifically, the mixup operation for each modality is defined as:

\begin{equation}
    \tilde{\mathbf{x}}_{i}^{m} = \lambda \mathbf{x}_{i}^{m} + (1 - \lambda) \mathbf{x}_{j}^{m}, \quad m\in {A,B,C}
\end{equation}

The labels for all modalities are mixed similarly, as follows:

\begin{equation}
\tilde{\mathbf{y}} = \lambda \mathbf{y}_i + (1 - \lambda) \mathbf{y}_j.
\end{equation}


\begin{figure*}
    \centering
    \includegraphics[width=1.0\linewidth]{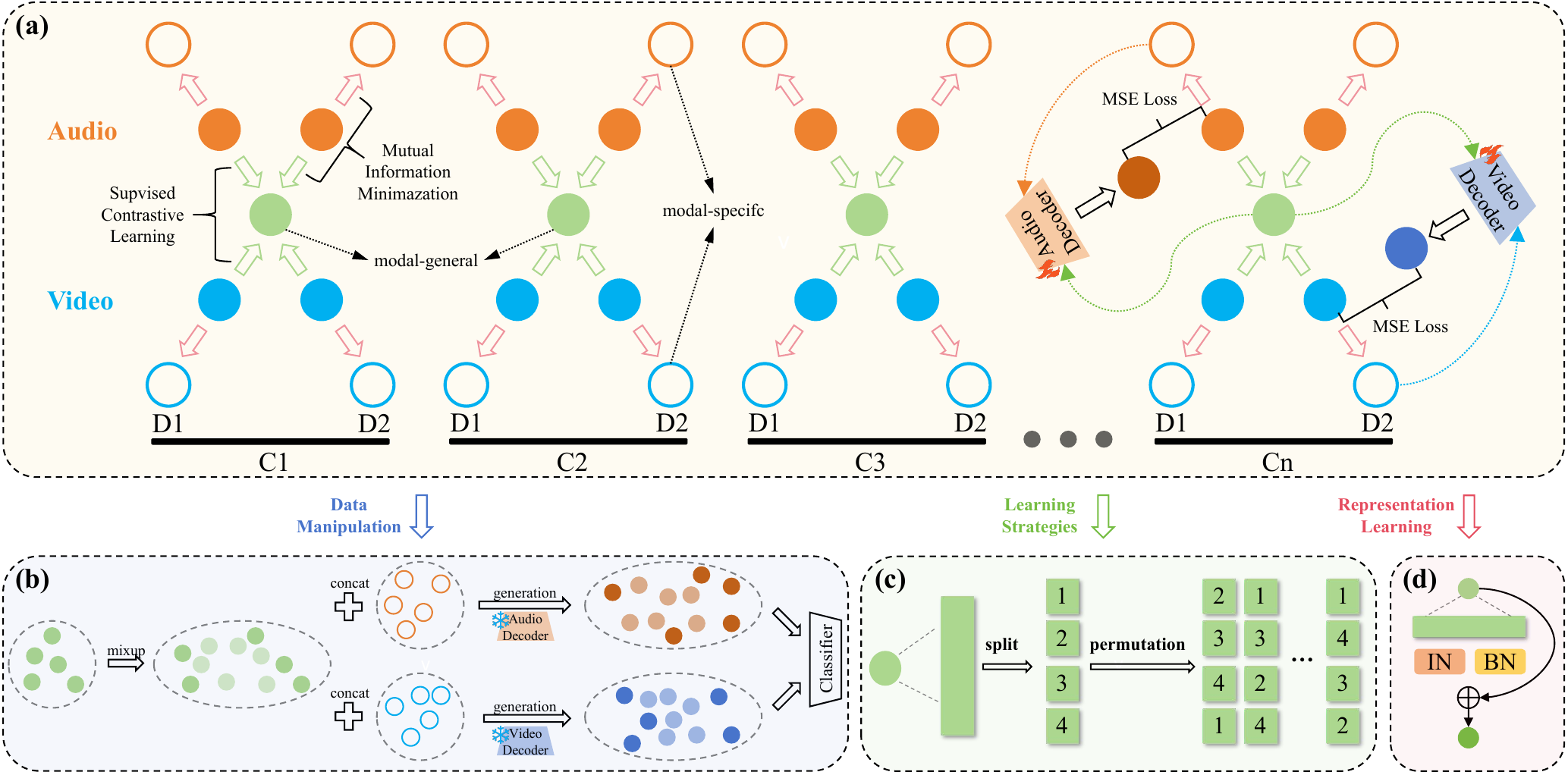}
    \vspace{-3mm}
    \caption{\textbf{Illustration of the URMMDG framework.} The framework first constructs a unified representation through supervised contrastive learning and disentanglement with multimodal data, and then applies single-modal DG methods on the unified representation. In \textbf{subfigure (a)}, \textbf{C} and \textbf{D} stand for \textbf{Class} and \textbf{Domain}, respectively, and the figure exemplifies our framework with \textbf{two domains} and \textbf{n classes}. \textbf{(b) Data Manipulation, (c) Learning Strategies, and (d) Representation Learning} are three different categories of DG methods, as described in~\cite{wang2022generalizing}.}
    \label{fig:framework}
    \vspace{-5mm}
\end{figure*}

\noindent
\textbf{JiGen: }
Given an embedding \( \mathbf{x}^m \) of the modality m, we split it into \( O \) equal-length segments, i.e., \( \mathbf{x}^m = [\mathbf{x}^m_{1}, \mathbf{x}^m_{2}, \dots, \mathbf{x}^m_{O}] \). These segments are then randomly shuffled to create different permutations. For example, if \( O = 3 \), one possible permutation could be \( \tilde{\mathbf{x}}^m_p = [\mathbf{x}^m_{2}, \mathbf{x}^m_{3}, \mathbf{x}^m_{1}] \). After shuffling, We generate \( P \) permutations from the \( O! \) possible configurations and assign a label to each permutation, where \( O! \) denotes the factorial of \( O \). This method is framed as an auxiliary classification problem, where each input \( \tilde{\mathbf{x}}^m_{o} \) is classified based on its permutation index \( o \). The loss is defined as the standard cross-entropy loss \( L_{\text{Jig}}(\mathcal{H}^m(\tilde{\mathbf{x}}^m_o), o) \), where \( \mathcal{H}^m \) represents the classifier that predicts the permutation index for modality \( m \) and $o \in \{1,...,P\}$. The total loss across all modalities, \( \sum_{m}L_{\text{Jig}} \), is then used as the baseline for the Jigsaw Puzzles task in this work.


\noindent
\textbf{IBN-Net: }
We adopt the classic IBN-a variant~\cite{pan2018two}, where Instance Normalization (IN) is applied to the first half of the channels, denoted as $\mathbf{x}_{\text{front}}$, in the shallow layers, while Batch Normalization (BN) is applied to the remaining half, $\mathbf{x}_{\text{back}}$. This architecture is well-suited for classification-related tasks.
\begin{equation}
    \hat{\mathbf{x}}^m = \text{IN}(\mathbf{x}^m_{\text{front}}) \oplus \text{BN}(\mathbf{x}^m_{\text{back}}) + \mathbf{x}^m
\end{equation}


\subsection{Supervised Contrastive Decoupling}
As shown in Figure~\ref{fig:framework}(a), the supervised contrastive disentanglement framework consists of two components: supervised contrastive learning aligns the general information across different modalities, while the disentanglement module decouples general and specific information within each modality. For each modality, we encode two representations: one representing the general information shared across multiple modalities and domains, and the other capturing specific information unique to each sample. Specifically, we denote the general information as $\mathbf{z}$ and the specific information as $\overline{\mathbf{z}}$. The category encoder and category-agnostic encoder are represented by $\Phi$ and $\Psi$, respectively, where $m \in \{\text{Video, Audio, Optical Flow}\}$.
\begin{equation}
\begin{split}
    \mathbf{z}^{m}_{i} &= \Phi^{m}(\mathbf{x}^{m}_{i}), 
    \overline{\mathbf{z}}^{m}_{i} = \Psi^{m}(\mathbf{x}^{m}_{i}), 
\end{split}
\label{equ1} 
\end{equation}


\noindent
\textbf{Multi-modal Supervised Contrastive Learning: }
To construct a unified multimodal representation, we employ supervised contrastive learning~\cite{khosla2020supervised, dong2023simmmdg} to achieve category-level semantic alignment across different modalities.
Let $i\in I\equiv\{1, ..., M \times N\}$ be the index of an arbitrary unimodal sample within a batch. We define $A(i)\equiv I\setminus\{i\}$, $P(i)\equiv\{p\in A(i):{\tilde{y}}_p={\tilde{y}}_i\}$ is the set of indices of all positives in the multiviewed batch
distinct from $i$, and $|P(i)|$ is its cardinality, the multi-modal supervised contrastive learning loss can be written as follows: 

\begin{equation}
  \mathcal{L}_{scl}
  =\sum_{i\in I}\frac{-1}{|P(i)|}\sum_{p\in P(i)}\log{\frac{\text{exp}\left(\boldsymbol{z}_i \cdot \boldsymbol{z}_p/\tau\right)}{\sum\limits_{a\in A(i)}\text{exp}\left(\boldsymbol{z}_i\cdot\boldsymbol{z}_a/\tau\right)}} ,
  \label{eqn:supervised_loss}
\end{equation}
where $\tau$ is the temperature parameter.

\noindent
\textbf{Mutual Information Minimization: }
CLUB~\cite{cheng2020club} optimizes an upper bound of mutual information to achieve effective information disentanglement. Given two variables $\mathbf{x}$ and $\mathbf{y}$, the objective function of CLUB is defined as:
\begin{equation}
\begin{split}
    I_{vCLUB}(\mathbf{x};\mathbf{y}):=\mathbb{E}_{p(\mathbf{x},\mathbf{y}}[\log q_{\theta}(\mathbf{y}|\mathbf{x})] 
    \\-\mathbb{E}_{p(\mathbf{x})}\mathbb{E}_{p(\mathbf{y})}[\log q_{\theta}(\mathbf{y}|\mathbf{x})].
\label{equ4} 
\end{split}
\end{equation}

We leverage it to optimize the upper bound of mutual information between the general information $\mathbf{z}^{m}_{i}$ and the specific information $\overline{\mathbf{z}}^{m}_{i}$, where $q_{\theta}$ is the variational approximation of the ground-truth posterior of $\mathbf{y}$ given $\mathbf{x}$ and can be parameterized by a network $\theta$.
\begin{equation}
\begin{split}
    L_{club}&=\frac{1}{N}\sum_{i=1}^N[\log q_{\theta}(\overline{\mathbf{z}}^{m}_{i}|\mathbf{z}^{m}_{i})\\&- \frac{1}{N}\sum_{j=1}^N\log q_{\theta}(\overline{\mathbf{z}}^{m}_{j}|\mathbf{z}^{m}_{i})]
\label{dec} 
\end{split}
\end{equation}

Additionally, we introduce a decoder to reconstruct the original input $x^{m}_{i}$ by merging the separated representations $\mathbf{z}^{m}_{i}$ and $\overline{\mathbf{z}}^{m}_{i}$, and compute the Mean Squared Error (MSE) loss to ensure semantic integrity after disentanglement. This also facilitates the subsequent Unified Representation Mixup in generating new multimodal mixup samples in a synchronized manner.
\begin{equation}
\begin{split}
    L_{rec}&=\|\mathbf{x}_{i}^{m} - D(\mathbf{z}_{i}^{m};\bar{\mathbf{z}}_{i}^{m})\|_2^2
\label{rec} 
\end{split}
\end{equation}

\noindent
\textbf{Unified Representation Mixup (UR-Mixup): }
As shown in Figure~\ref{fig:framework}(b), we apply Mixup to the general information $\mathbf{z}^{m}_{i}$, which encapsulates the core category semantics, making it a suitable target for Mixup-based data augmentation. By leveraging the unified representation, we achieve synchronized multimodal Mixup, enhancing model generalization. 
\begin{equation}
    \tilde{\mathbf{z}}_{i}^{m} = \lambda \mathbf{z}_{i}^{m} + (1 - \lambda) \mathbf{z}_{j}^{m}, \tilde{\mathbf{y}} = \lambda \mathbf{y}_i + (1 - \lambda) \mathbf{y}_j
\end{equation}

The mixed augmented category semantics from Mixup is then combined with the specific information and fed into the decoder $D$, generating domain-aware features with synchronized multimodal augmentation:
\begin{equation}
    \tilde{\mathbf{x}}_{i}^{m} = D(\tilde{\mathbf{z}}_{i}^{m},\bar{\mathbf{z}}_{i}^{m}))
\end{equation}
The generated new data is then used for classification training along with the corresponding augmented soft labels $\tilde{\mathbf{y}}$, optimized using cross-entropy loss.




\noindent
\textbf{Unified Representation JiGen (UR-JiGen): }
Previous works~\cite{noroozi2016unsupervised,carlucci2019domain} have employed Jigsaw puzzles as a self-supervised task to learn visual representations by reconstructing an image from its shuffled parts. Inspired by this, UR-JiGen utilizes general information ${\mathbf{z}}^{m}_{i}$ to enable synchronized self-supervised learning across multiple modalities.

We divide the general information into $O$ equal-length parts: $z^m = [z^m_1, z^m_2, \dots, z^m_O]$. These segments are randomly selected across modalities to form $z^r = [z^{m_1}_1, z^{m_2}_2, \dots, z^{m_o}_O]$, where \( m_1, m_2, m_o \) can be the same or different modalities. One possible permutation is $\tilde{z}^o = [z^{m_2}_2, z^{m_n}_O, \dots, z^{m_1}_1]$. The $O$ segments are then shuffled to generate different permutations, resulting in $O!$ possible combinations. From these, we select $P$ permutations and assign an index to each as a label.

We introduce an auxiliary classification task for each sample instance, the loss is defined as the standard cross-entropy loss \( L_{\text{Jig}}(\mathcal{H}(\tilde{\mathbf{z}}^m_o), o) \), where \( \mathcal{H} \) represents the classifier that predicts the permutation index and $o \in \{1,...,P\}$. 

\noindent
\textbf{Unified Representation IBN-Net (UR-IBN): }
Unified Representation IBN-Net applies IBN-a directly to the unified representation, where Instance Normalization (IN) is applied to half of the dimensions, Batch Normalization (BN) to the other half, followed by a residual connection.
\begin{equation}
    \hat{\mathbf{z}}^m = \text{IN}(\mathbf{z}^m_{\text{front}}) \oplus \text{BN}(\mathbf{z}^m_{\text{back}}) + \mathbf{z}^m
\end{equation}

\subsection{Final Loss}
The final loss functions for UR-Mixup and UR-IBN are formulated as follows:
\begin{equation}
    L = \alpha_{1} L_{\text{cls}} + \alpha_{2} L_{\text{scl}} + \alpha_{3}L_{\text{club}} + \alpha_{4}L_{\text{rec}},
\end{equation}
UR-JiGen additionally incorporates the loss term \( L_{jig} \), where \( L_{\text{cls}} \) is the cross-entropy loss for classification. The hyperparameters \( \alpha_1 \), \( \alpha_2 \), \( \alpha_3 \), and \( \alpha_4 \) control the relative importance of classification, supervised contrastive learning, mutual information minimization, and reconstruction, respectively, the weight for \( L_{jig} \) is set to 1.


%% file: sec/4_experiment.tex
\section{Experiment}
\label{sec:experiment}


%


\subsection{Experimental Setting}
\textbf{Dataset: }We evaluate our method on two benchmarks: EPIC-Kitchens~\cite{damen2018scaling} and Human-Animal-Cartoon (HAC)~\cite{dong2023simmmdg}, both of which comprise three modalities: video, audio, and optical flow. EPIC-Kitchens is divided into three domains: D1, D2, and D3, while HAC consists of three domains: humans (H), animals (A), and cartoon figures (C). The experimental setup follows the configurations of SimMMDG~\cite{dong2023simmmdg} and CMRF~\cite{CMRF}. For details on implementation and hyperparameter analysis, please refer to the supplementary material.


\noindent
\textbf{Baseline: }We compare our model with the following baselines:  
1) \textbf{Base}: vanilla multi-modal joint training without any domain generalization strategies.  
2) \textbf{RNA-Net}~\cite{planamente2022domain}, \textbf{SimMMDG}~\cite{dong2023simmmdg}, and \textbf{CMRF}~\cite{CMRF}: models specifically designed for MMDG and representing the previous state-of-the-art.  
3) \textbf{Mixup}~\cite{zhang2017mixup}, \textbf{JiGen}~\cite{carlucci2019domain}, and \textbf{IBN-Net}~\cite{pan2018two}: DG methods directly applied to the MMDG setting.




\input{table/epic_hac_mmdg_ms}
\input{table/epic_hac_mmdg_ss}

\subsection{Performance} 
For all tables, rows with a gray background indicate DG methods directly transferred to MMDG. \textbf{Bold} and \underline{underlined} numbers represent the best and second-best results, respectively, and all values in the tables are reported as top-1 accuracy (\%).

\noindent
\textbf{Multi-modal multi-source DG: }Table~\ref{tab:epic_hac_mmdg_ms} presents the experimental results of our method and all baseline models on the EPIC-Kitchens and HAC datasets under the multi-modal multi-source domain generalization setting, where models are trained on multiple source domains and tested on one target domain. We conduct experiments using different modality combinations, including any two modalities as well as all three modalities jointly, to evaluate the generalization capability of our approach. It can be observed that directly transferring DG methods to MMDG provides some improvement over the Base model but still lags significantly behind MMDG state-of-the-art methods such as SimMMDG~\cite{dong2023simmmdg} and CMRF~\cite{CMRF}. In contrast, our proposed approach significantly enhances the performance of DG methods in the MMDG setting, achieving an average improvement of 1.41\% to 3.49\% on EPIC-Kitchens and an even more notable improvement on HAC, where Mixup gains up to 5.95\%. Additionally, UR-JiGen and UR-Mixup demonstrate performance comparable to or even surpassing state-of-the-art methods, without requiring complex, MMDG-specific architectural modifications.


\noindent
\textbf{Multi-modal single-source DG: }Our method does not rely on any domain labels, making it applicable to multi-modal single-source domain generalization, where the model is trained on a single source domain and tested across multiple target domains. Table~\ref{tab:epic_hac_mmdg_ss} presents the experimental results obtained using all three modalities for training. Compared to previous state-of-the-art approaches, UR-JiGen and UR-Mixup continue to demonstrate highly competitive performance, while UR-IBN achieves a notable improvement over IBN. Specifically, UR-JiGen outperforms SimMMDG on average across both EPIC-Kitchens and HAC, achieving results comparable to CMRF. Meanwhile, UR-Mixup surpasses CMRF significantly, with scores of 60.84 vs. 60.12 on EPIC-Kitchens and 64.91 vs. 64.09 on HAC. This further validates the effectiveness of our proposed unified representation as a bridge to facilitate the transfer of DG methods to MMDG. Additional results for training with bimodal combinations are provided in the supplementary material.


\noindent
\textbf{Uni-modal performance in MMDG: }
In Table~\ref{tab:modal_compete} of the Introduction, we discussed the suboptimal performance when directly transferring DG methods to MMDG. Two key issues arise: (1) DG methods under multi-modal training yield significantly smaller improvements compared to their single-modal training; (2) models trained jointly on multiple modalities perform worse in single-modal testing than those trained solely on a single modality. This indicates an optimization misalignment when applying DG methods across modalities, leading to modality competition~\cite{huang2022modality}. Previous work, such as CMRF~\cite{CMRF}, has highlighted this issue but has not proposed a general solution.  

As shown in Table~\ref{tab:hac_compete}, we further explore this phenomenon through extensive experiments on three different bimodal combinations (Base, JiGen, and Mixup exhibit lower single-modal test performance when trained jointly on multiple modalities compared to single-modal training. Moreover, JiGen and Mixup provide significantly greater improvements in single-modal settings than in multi-modal joint training), reaffirming the conclusions drawn from Table~\ref{tab:modal_compete}. Additionally, our models trained in a multi-modal setting not only achieve substantial improvements in multi-modal testing but also surpass single-modal DG methods in single-modal training (e.g., in UR-Mixup with Video-Audio training, the single-modal test performance for Video reaches 71.64, surpassing the 71.12 achieved by Mixup in single-modal training). This demonstrates that our approach effectively leverages cross-modal knowledge. More experimental results on EPIC-Kitchens are provided in the supplementary material.

\input{table/hac_compete}

\begin{figure}[h]
    \centering
    \includegraphics[width=0.87\linewidth]{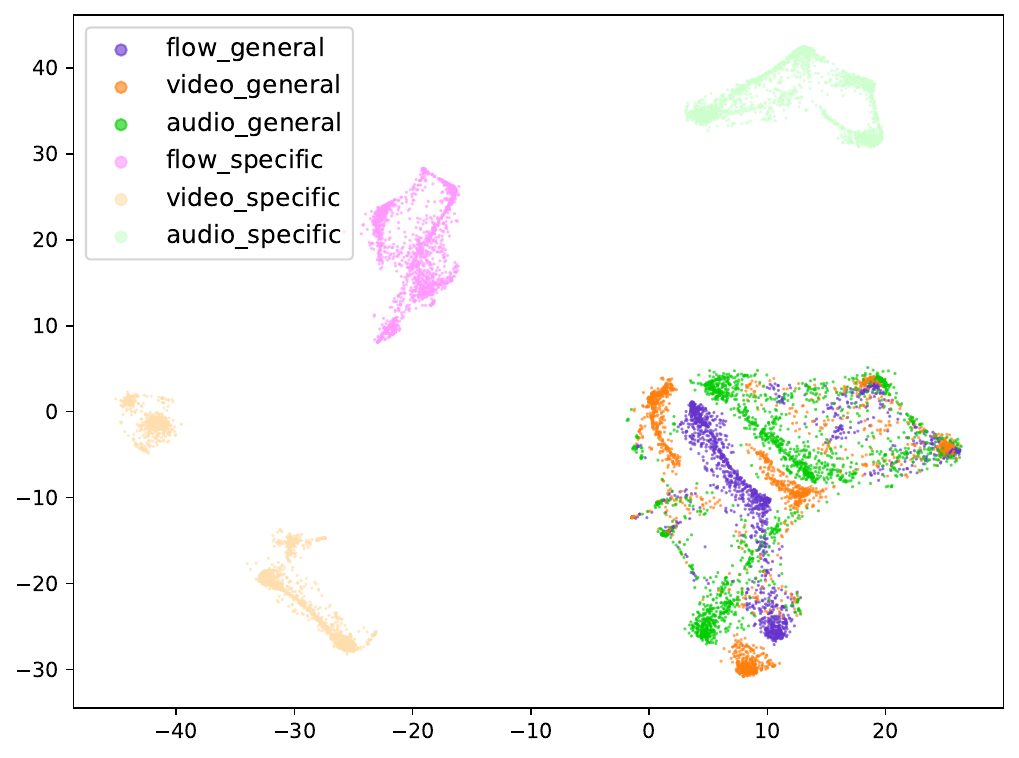}
    \caption{Visualization of the learned embeddings using t-SNE (D2, D3 → D1 in EPIC-Kitchens for multi-modal multi-source DG).}
    \label{fig:representation_visualization}
\end{figure}

\noindent
\textbf{Representation Visualization: }
We visualize the general and specific information for the three modalities, as shown in Figure~\ref{fig:representation_visualization}. The general information across different modalities is closely clustered together, while the specific information is distinctly separated from the general representations. This demonstrates the effectiveness of our proposed unified representation construction. Although the general distributions of different modalities are not perfectly identical, they exhibit a similar overall trend, further validating the consistency of our approach. We also provide visualizations demonstrating the disentanglement of source and target domains for each of the three modalities, further highlighting the effectiveness of our modality disentanglement. Please refer to the supplementary material for details.


\subsection{Ablation Studies}

\textbf{Ablation on Unified Representation:}As shown in Table~\ref{tab:ablation_UR}, comparing the second and third rows reveals that supervised contrastive learning outperforms unsupervised contrastive learning. This is because supervised contrastive learning leverages label information, enabling the model to construct a unified representation space centered around categories. Additionally, the remaining results demonstrate that CLUB~\cite{cheng2020club} is more effective in disentangling mutual information compared to using a simple MSE loss. These findings validate the effectiveness of our proposed approach for constructing a unified representation.
\input{table/ablation_UR}





\noindent
\textbf{Ablation on UR-JiGen and UR-Mixup:}
We conducted further ablation studies on UR-Mixup and UR-JiGen. For UR-JiGen, we evaluated four different configurations of the Jigsaw number \( P \), with \( P=256 \) achieving the best performance. For UR-Mixup, we tested different mixing strategies and found that Rand Mix yielded superior results.
\input{table/ablation_mixup_jigen}


%



%% file: table/epic_hac_mmdg_ms.tex
\begin{table*}[t!]
\centering
\resizebox{\linewidth}{!}{
\begin{threeparttable}
\begin{tabular}{lcccccccccccc}
\toprule
& \multicolumn{3}{c}{\textbf{Modality}} & \multicolumn{4}{c}{\textbf{EPIC-Kitchens dataset}}& \multicolumn{4}{c}{\textbf{HAC dataset}}\\
\cmidrule(lr){2-4} \cmidrule(lr){5-8} \cmidrule(lr){9-12} 
\textbf{Method} & Video & Audio & Flow & D2, D3 $\rightarrow$ D1 & D1, D3 $\rightarrow$ D2 & D1, D2 $\rightarrow$ D3  & \textit{Avg} & A, C $\rightarrow$ H & H, C $\rightarrow$ A & H, A $\rightarrow$ C  & \textit{Avg}\\


\midrule

Base & $\checkmark$& $\checkmark$&    & 54.94 & 62.26 & 61.70 &  59.63& 69.92  & 69.32  &  50.09 & 63.11 \\
RNA-Net~\cite{planamente2022domain} & $\checkmark$& $\checkmark$&  & 55.37 & 64.20 & 62.25  &60.61 & 67.45  & 68.32  &  54.78 &  63.52\\
SimMMDG~\cite{dong2023simmmdg}& $\checkmark$& $\checkmark$&   & 		\underline{57.24} & 65.07 & 63.55 & 61.95& 	72.75 & 76.14 & 54.59 & 67.83 \\
CMRF~\cite{CMRF}& $\checkmark$& $\checkmark$&   & 		56.55 & \underline{68.13} & \underline{67.04} & \underline{63.91} & 		76.45 & \textbf{82.39}& \underline{56.88} & \underline{71.91} \\

\rowcolor{gray!30} 
IBN~\cite{pan2018two} & $\checkmark$& $\checkmark$&    & 55.28 & 63.15 & 61.58 &  60.00 & 69.51  & 71.67  &  52.64 & 64.61 \\
\rowcolor{gray!30} 
JiGen~\cite{carlucci2019domain} & $\checkmark$& $\checkmark$&    & 56.21 & 64.45 & 62.58 & 61.08 & 70.21  & 73.21  &  53.98 & 65.80 \\
\rowcolor{gray!30} 
Mixup~\cite{zhang2017mixup} & $\checkmark$& $\checkmark$&    & 57.11 & 63.68 & 63.05 & 61.28 & 71.38  & 73.29  &  54.57 & 66.41 \\

UR-IBN (ours) & $\checkmark$& $\checkmark$&    & 56.73 & 64.41 & 63.08 &  61.41 & 69.98  & 74.52  & 53.97 & 66.16 \\
UR-JiGen (ours) & $\checkmark$& $\checkmark$&    & \textbf{57.62} & 67.40 & 65.88 & 63.63 & \underline{76.97}  & 81.35  &  55.87 & 71.40 \\
UR-Mixup (ours) & $\checkmark$& $\checkmark$&    & 56.99 & \textbf{68.85} & \textbf{68.46} & \textbf{64.77} & \textbf{77.57}  & \underline{82.04}  & \textbf{57.48} & \textbf{72.36} \\

\midrule

Base & $\checkmark$& &$\checkmark$ & 55.86 & 67.47 & 59.34 &60.89 & 72.83  &  77.84 &43.58 &  64.75 \\
RNA-Net~\cite{planamente2022domain} & $\checkmark$& & $\checkmark$& 54.21 & 64.80 & 59.31 &59.44 &  74.56 &75.39  &44.90 &  64.95\\
SimMMDG~\cite{dong2023simmmdg}& $\checkmark$& & $\checkmark$& 57.03 & 66.67 & 63.86 & 62.82& 77.90 & 78.98 & \textbf{57.80} & 71.56 \\
CMRF~\cite{CMRF}& $\checkmark$& &$\checkmark$   & 	\textbf{65.28} & 67.87& \underline{64.89}& \underline{66.01}& 	\underline{81.16}& \textbf{81.25} & 55.50 & \underline{72.64}\\

\rowcolor{gray!30} 
IBN~\cite{pan2018two} & $\checkmark$& &$\checkmark$    & 55.82 & 68.13 & 59.91 & 61.29 & 73.28  &77.63  &  44.64 & 65.18 \\
\rowcolor{gray!30} 
JiGen~\cite{carlucci2019domain} & $\checkmark$& &$\checkmark$    & 58.67 & 66.62 & 60.08 & 61.79 & 75.61 & 76.29  &  52.47 & 68.12 \\
\rowcolor{gray!30} 
Mixup~\cite{zhang2017mixup} & $\checkmark$& &$\checkmark$    & 59.85 & 66.22 & 60.17 & 62.08 & 78.89  & 76.02  &  51.88 & 68.93 \\

UR-IBN (ours) & $\checkmark$& &$\checkmark$    &57.20&\underline{68.72}&60.72&62.21&75.35&78.81&47.22&67.13 \\
UR-JiGen (ours) & $\checkmark$& &$\checkmark$    &62.17&66.37&63.90&64.15&80.52&79.15&\underline{57.25}&72.31 \\
UR-Mixup (ours) & $\checkmark$& &$\checkmark$    & \underline{64.85} & \textbf{68.84} & \textbf{65.57} & \textbf{66.42} & \textbf{82.17}  & \underline{80.43}  & 56.85 & \textbf{73.15} \\

\midrule

Base & & $\checkmark$&$\checkmark$ & 49.42 & 55.60 & 54.41 & 53.14 & 52.89 & 55.11 &40.92 & 49.64 \\
RNA-Net~\cite{planamente2022domain} & & $\checkmark$&$\checkmark$ & 50.89 & 54.24 & 55.90 & 53.68  & 53.11 & 59.32& 43.82&  52.08\\
SimMMDG~\cite{dong2023simmmdg}& & $\checkmark$&$\checkmark$ & 55.86 & 64.60 & 59.34 & 59.93 & 57.88 & 60.79 & 48.62 & 55.76\\
CMRF~\cite{CMRF}& &$\checkmark$ &$\checkmark$   & 	\underline{57.24} & 64.94 & \textbf{66.12} & \underline{62.76} & 	\underline{59.06} & \underline{61.79} & \underline{55.04} &58.49 \\

\rowcolor{gray!30} 
IBN~\cite{pan2018two} & &$\checkmark$ &$\checkmark$    & 49.89 & 54.61 & 56.85 & 53.78 & 54.62  & 57.74  & 44.20 & 52.19 \\
\rowcolor{gray!30} 
JiGen~\cite{carlucci2019domain} & &$\checkmark$ &$\checkmark$    & 53.57 & 61.79 & 61.42 & 58.93 & 55.44  & 60.58  & 47.69 & 54.57 \\
\rowcolor{gray!30} 
Mixup~\cite{zhang2017mixup} & &$\checkmark$ &$\checkmark$    & 52.85 & 61.28 & 60.83 & 58.32& 55.81  & 58.83  &  50.64 & 55.09 \\

UR-IBN (ours) & &$\checkmark$ &$\checkmark$    & 55.23 & 63.26 & 60.17 & 59.55& 56.89  & 61.68  & 49.18 & 55.92 \\
UR-JiGen (ours) & &$\checkmark$ &$\checkmark$    & \textbf{58.40} & \underline{65.72} & \underline{65.84} & \textbf{63.32} & 58.63 & \textbf{62.25}  & 54.84 & \underline{58.56} \\
UR-Mixup (ours) & &$\checkmark$ &$\checkmark$    & 56.87 & \textbf{66.41} & 64.62 & 62.63 & \textbf{59.44}  &61.63  & \textbf{55.46} & \textbf{58.84} \\

\midrule

Base & $\checkmark$& $\checkmark$& $\checkmark$ & 54.71 & 67.20 & 61.70 & 61.20& 70.29& 71.25 &53.57 &65.07  \\
RNA-Net~\cite{planamente2022domain} & $\checkmark$& $\checkmark$& $\checkmark$ &56.25 & 63.47 & 59.72 & 59.81 &  71.89& 70.88 &54.58& 65.78 \\
SimMMDG~\cite{dong2023simmmdg} & $\checkmark$& $\checkmark$& $\checkmark$ & \underline{62.08} & 66.13 & 64.40& 64.20 & 76.27&77.70&56.42& 70.13 \\
CMRF~\cite{CMRF}&$\checkmark$ &$\checkmark$ &$\checkmark$   & 		61.84 & 70.13 & \underline{70.12} & \underline{67.36}& 	78.26&\underline{79.54} & \underline{60.09} &72.44 \\

\rowcolor{gray!30} 
IBN~\cite{pan2018two} &$\checkmark$ &$\checkmark$ &$\checkmark$    & 55.76 & 64.75 & 62.43 & 60.98 & 69.26  & 72.62  & 54.28 & 65.39 \\
\rowcolor{gray!30} 
JiGen~\cite{carlucci2019domain} &$\checkmark$ &$\checkmark$ &$\checkmark$    & 57.37 & 67.73 & 63.34 & 62.81& 72.45  & 74.83  & 55.26 & 67.49 \\
\rowcolor{gray!30} 
Mixup~\cite{zhang2017mixup} &$\checkmark$ &$\checkmark$ &$\checkmark$    & 57.95 & 67.95 & 64.37 & 63.42& 73.37  & 74.49  &  55.48 & 67.78 \\

UR-IBN (ours) &$\checkmark$ &$\checkmark$ &$\checkmark$    & 59.62 & 67.93 & 66.61 &  64.72 & 73.29  & 75.18  & 57.24 & 68.57 \\
UR-JiGen (ours) &$\checkmark$ &$\checkmark$ &$\checkmark$    & \textbf{62.20} &\textbf{71.14} & 67.78 & 67.04 & \underline{78.47}  & 79.26  &  59.83 & \underline{72.52} \\
UR-Mixup (ours) &$\checkmark$ &$\checkmark$ &$\checkmark$    & 61.72 & \underline{70.89} & \textbf{70.76} &  \textbf{67.79} & \textbf{78.72}  & \textbf{80.63}  &  \textbf{60.85} & \textbf{73.40} \\

\bottomrule
\end{tabular}

\end{threeparttable}
}
\caption{Multi-modal \textbf{multi-source} DG with different modalities on EPIC-Kitchens and HAC datasets.}
\label{tab:epic_hac_mmdg_ms}
\end{table*}

%% file: table/epic_hac_mmdg_ss.tex
\begin{table*}[t!]
\centering
\resizebox{\linewidth}{!}{
\begin{threeparttable}
\begin{tabular}{lccccccccccccccc}
\toprule
& \multicolumn{6}{c}{\textbf{EPIC-Kitchens dataset}}&& \multicolumn{6}{c}{\textbf{HAC dataset}}\\
\cmidrule(lr){2-7} \cmidrule(lr){9-14}
\qquad\qquad\; Source: & \multicolumn{2}{c}{D1}& \multicolumn{2}{c}{D2}& \multicolumn{2}{c}{D3}& & \multicolumn{2}{c}{H}& \multicolumn{2}{c}{A}& \multicolumn{2}{c}{C}\\
\cmidrule(lr){2-3} \cmidrule(lr){4-5} \cmidrule(lr){6-7} \cmidrule(lr){9-10} \cmidrule(lr){11-12} \cmidrule(lr){13-14} 

\textbf{Method} \quad Target: &  D2 &  D3 &  D1 &  D3& D1&  D2  & \textit{Avg}& A & C & H & C& H& A  & \textit{Avg}\\
\midrule
Base  &56.80& 53.08& 47.36 &59.65 &55.63 &56.93 &54.91 &64.20 &39.45&64.85& 52.29& 57.97& 65.90& 57.44\\
RNA-Net~\cite{planamente2022domain}  & 50.32 &51.27 &48.90 &61.34 &53.76 &55.89 &53.58 &62.35 &43.24 &64.21 &53.46 &55.37 &66.82 &57.57\\
SimMMDG~\cite{dong2023simmmdg}& 54.13 &\textbf{57.90} &50.57 &63.04& \textbf{60.69} &64.27& 58.43 &64.77 &39.44 &71.38 &50.46& 60.14& 70.77& 59.49 \\
CMRF~\cite{CMRF} &  \textbf{60.80}& 56.78& \underline{55.17}& \underline{64.99} & 57.24& \underline{65.73}& \underline{60.12}& \underline{68.75}& \textbf{46.33} &73.55 &\underline{58.26} &65.22 &\underline{72.46} &\underline{64.09} \\
\rowcolor{gray!30} 
IBN~\cite{pan2018two} &  55.67& 54.10&  48.49&  60.57   & 54.99      & 57.08  & 55.15 & 64.89 & 41.22&65.75 & 52.90 &  56.82 & 66.91  & 58.08 \\
\rowcolor{gray!30} 
JiGen~\cite{carlucci2019domain} &  57.37& 55.23&  50.02&  61.14   & 54.63      & 57.29  & 55.95 & 65.63 & 41.61&67.36 & 51.84 & 56.72 & 67.86  & 58.50 \\
\rowcolor{gray!30} 
Mixup~\cite{zhang2017mixup} &  57.83& 54.49&  53.65&  62.71   & 56.94      & 57.16  & 57.13 & 64.49 & 41.68 &66.49 & 53.19 &  58.62 & 67.74  & 58.70 \\

UR-IBN (ours) &  55.89& 56.13&  51.33&  61.69   & 54.83      & 57.62&56.25& 66.13 & 43.24& 65.27 & 54.08 &  59.94 & 69.39  & 59.67 \\
UR-JiGen (ours) &  59.63& 56.21&  54.35&  63.79   & 58.26  & 64.31  & 59.43    &67.27 & 43.38&\textbf{74.82} & \textbf{58.79} &  \underline{66.57} & 71.31  & 63.69 \\
UR-Mixup (ours) &  \underline{60.68}& \underline{57.16}&  \textbf{55.74}&  \textbf{65.11}   & \underline{59.63} & \textbf{66.74}  & \textbf{60.84} & \textbf{69.43} & \underline{45.85} &\underline{74.52} & 58.24 &  \textbf{67.96} & \textbf{73.48}  & \textbf{64.91} \\

\bottomrule
\end{tabular}

\end{threeparttable}
}
\caption{Multi-modal \textbf{single-source} DG on EPIC-Kitchens and HAC datasets using video, flow and audio three modalities.}
\label{tab:epic_hac_mmdg_ss}
\end{table*}

%% file: table/hac_compete.tex
\begin{table*}[]
    \centering
    \resizebox{1.0\linewidth}{!}{
    \begin{tabular}{lccccccccc}
\toprule
 \textbf{Method} & Video & Audio & Video-Audio  & Video & Flow & Video-Flow & Audio & Flow & Audio-Flow\\
\cmidrule(lr){1-1} \cmidrule(lr){2-4} \cmidrule(lr){5-7}  \cmidrule(lr){8-10}
Base (M1) &  68.07 & - & - & 68.07 & - & -& 32.81 & - & -\\
Base (M2)&  - & 32.81 & - & - & 53.20 & -& - & 53.20 & -\\
Base (MM) &  67.60& 31.24& 63.11& 64.29 & 48.64 & 64.75&30.28 & 50.39 & 49.64\\
\midrule
JiGen (M1) & 70.41 & - & - & 70.41 & - & -& 35.26 & - & -\\
JiGen (M2) &  - & 35.26 & - & - & 54.36 & -& - & 54.36 & -\\
JiGen (MM) & 63.42& 32.08 & 65.80 & 66.28 & 51.87 & 68.12& 31.14 & 51.50 & 54.57\\
\midrule
Mixup (M1) &  \underline{71.12}& - & - &71.12& - & -& \underline{35.83} & - & -\\
Mixup (M2) &  - & \underline{35.83} & - & -& \textbf{54.84} & -& - & \underline{54.84} & -\\
Mixup (MM) &68.22 & 31.64 & 66.41 & 66.71 & 52.48 & 68.93& 32.43 & 51.76 & 55.09\\
\midrule
UR-JiGen (ours) &  70.79& 35.14 & \underline{71.40} & \underline{69.89} & \underline{54.74} & \underline{72.31}& 35.69 & 53.97 & \underline{58.56}\\
UR-Mixup (ours) &  \textbf{71.64} & \textbf{36.21} & \textbf{72.36} & \textbf{71.35} & 54.60 & \textbf{73.15} &\textbf{36.27} & \textbf{55.08} & \textbf{58.84}\\
\bottomrule
    \end{tabular}}
    \caption{The average results of uni-modal performance comparison under multi-modal multi-source DG on HAC with 3 different modality combinations. M1, M2, and MM denote training settings where the data correspond to the first and second single-modal cases, and the multi-modal case, respectively, following the column header order.
}
    \label{tab:hac_compete}
\end{table*}

%% file: table/ablation_UR.tex
\setlength{\tabcolsep}{2pt}
\begin{table}
\centering
\begin{adjustbox}{width=0.98\linewidth}

\begin{tabular}{cccc|ccccc}
\toprule
UCL & SCL & MID & CID &  D2, D3 $\rightarrow$ D1 & D1, D3 $\rightarrow$ D2 & D1, D2 $\rightarrow$ D3  & \textit{Mean}\\
\midrule

& & &  & 54.94& 62.26 &61.70 &59.63 \\
$\checkmark$ & & &  & 55.72 & 64.73 & 63.69 & 61.38 \\
 & $\checkmark$& &  & 56.24 & 65.38 & 65.07 & 62.23 \\
$\checkmark$ & &$\checkmark$ &  & 54.12 & 67.47 & 66.48 & 62.69 \\
&$\checkmark$&$\checkmark$ &  & 56.26 & 66.82 & \underline{68.37} & 63.82 \\
$\checkmark$ & & &$\checkmark$  & \underline{56.45} & \underline{67.69} & 68.22 & \underline{64.12} \\
&$\checkmark$& &$\checkmark$  & \textbf{56.99} & \textbf{68.85} & \textbf{68.46} & \textbf{64.77} \\

\bottomrule
\end{tabular}
\end{adjustbox}

\caption{Based Mixup ablations of unified representations on EPIC-Kitchens with video and audio data. UCL: Unsupervised Contrastive Learning, SCL: Supervised Contrastive Learning, MID: MSE Information Decoupling, CID: CLUB Information Decoupling}

\label{tab:ablation_UR}
\end{table}

%% file: table/ablation_mixup_jigen.tex
\setlength{\tabcolsep}{2pt}
\begin{table}
\centering
\begin{adjustbox}{width=0.98\linewidth}

\begin{tabular}{lcccc}
\toprule
Method &  D2, D3 $\rightarrow$ D1 & D1, D3 $\rightarrow$ D2 & D1, D2 $\rightarrow$ D3  & \textit{Mean}\\
\midrule
UR-JiGen (128)& 56.03 & 66.36 & 65.97 & 62.79 \\
UR-JiGen (256)& \textbf{57.62} &67.40&65.88 & \underline{63.63} \\
UR-JiGen (384)& 56.46 & 67.14 & 65.42 & 63.01 \\
UR-JiGen (512)& 56.21 & 66.58 & 66.56 & 63.12 \\
\midrule
UR-Mixup (Fixed Mix)& 55.37 & \underline{68.23} & \underline{66.85} & 63.48 \\
UR-Mixup (Rand Mix) & \underline{56.99}& \textbf{68.85} & \textbf{68.46} & \textbf{64.77} \\
\bottomrule
\end{tabular}
\end{adjustbox}

\caption{Based Mixup ablations of UR-Mixup and UR-JiGen on
EPIC-Kitchens with video and audio data. Fixed Mix: interpolations with fixed mixing ratio (0.5). The number following UR-JiGen represents the Jigsaw number $P$.}

\label{tab:ablation_mixup_jigen}
\end{table}

%% file: sec/5_conclusion.tex
\section{Conclusion}
\label{sec:conclusion}
In this paper, we identify modality discrepancies as a key challenge that limits the transfer of single-modal DG methods to MMDG, causing conflicts in learning dynamics. To address this, we propose a novel framework that constructs a highly aligned unified multimodal representation space, reformulating MMDG as a DG problem to mitigate modality asynchrony and enhance cross-modal consistency. Through extensive analysis and empirical validation, we demonstrate the effectiveness of our approach. Experimental results across multi-modal multi-source DG, multi-modal single-source DG, and uni-modal generalization in MMDG show that our method significantly outperforms conventional DG adaptations. Notably, UR-JiGen achieves performance on par with state-of-the-art models, while UR-Mixup surpasses them, highlighting the robustness and efficacy of our unified representation learning paradigm. These findings establish a new perspective on bridging DG and MMDG, offering a promising direction for future research.

%% file: sec/X_supplyment.tex
\clearpage
\setcounter{page}{1}
\maketitlesupplementary


\section{Inplementation Details}
To ensure fair experimental comparisons, we adopt the same modality backbones as SimMMDG~\cite{dong2023simmmdg} and CMRF~\cite{CMRF}, with our experimental setup based on the MMAction2 toolkit~\cite{contributors2020openmmlab}. The feature dimensions for video, audio, and optical flow are 2304, 512, and 2048, respectively. In constructing the unified representation, both the general information and specific information maintain a consistent dimension of 512 across all modalities. Each modality's general encoder and specific encoder consist of a two-layer MLP with an input dimension matching the respective modality's feature dimension (2304, 512, or 2048), a hidden layer of size 2048, and an output dimension of 512. The scalar temperature parameter $\tau$ is set to 0.1.

For modality-specific encoders, we use the SlowFast network's slow-only pathway for optical flow encoding, initialized with Kinetics-400 pre-trained weights, the SlowFast network~\cite{feichtenhofer2019slowfast} for visual encoding, also initialized with Kinetics-400 pre-trained weights~\cite{kay2017kinetics}, and ResNet-18~\cite{he2016deep} for audio encoding, initialized with weights from the VGGSound pre-trained checkpoint~\cite{chen2020vggsound}. The hyperparameters \( \alpha_1, \alpha_2, \alpha_3, \alpha_4 \) are set to 1.0, 2.0, 2.0, and 1.0, respectively. For UR-Mixup, the Beta distribution parameter \( \alpha \) is set to 0.2, while for UR-JiGen, the Jigsaw number \( P \) is set to 256.

All experiments are conducted on an NVIDIA GeForce RTX 4090 GPU, with training performed for 20 epochs. UR-IBN and UR-Mixup require approximately 3 hours, while UR-JiGen takes around 3.5 hours. In our implementation, we initially experimented with a warm-start approach, where training was conducted in two phases: first, learning the unified representation, followed by applying the DG method. However, experimental results indicated that warm-start had no significant impact. Consequently, we adopted a more efficient end-to-end training strategy, where the construction of the unified representation and the application of the DG method occur simultaneously.



\section{More Experiments}
\noindent
\textbf{More Experiments about Multi-modal single-source DG: }
As shown in Table~\ref{tab:sup_epic_hac_singlesource}, directly transferring DG methods~\cite{pan2018two,carlucci2019domain,zhang2017mixup} to MMDG results in significantly inferior performance compared to models specifically designed for MMDG~\cite{dong2023simmmdg,CMRF}. In contrast, our proposed approach substantially improves their performance in the MMDG setting. Notably, UR-JiGen and UR-Mixup demonstrate competitive results against previous state-of-the-art models, further validating the effectiveness of our method.
\input{table/sup_epic_hac_singlesource}

\noindent
\textbf{More Experiments about Uni-modal performance in MMDG: }
As shown in Table~\ref{tab:epic_compete}, the results are consistent with those in Table~\ref{tab:epic_compete}, further reinforcing our previous findings.
\input{table/epic_compete}


\begin{figure*}
    \centering
    \includegraphics[width=1\linewidth]{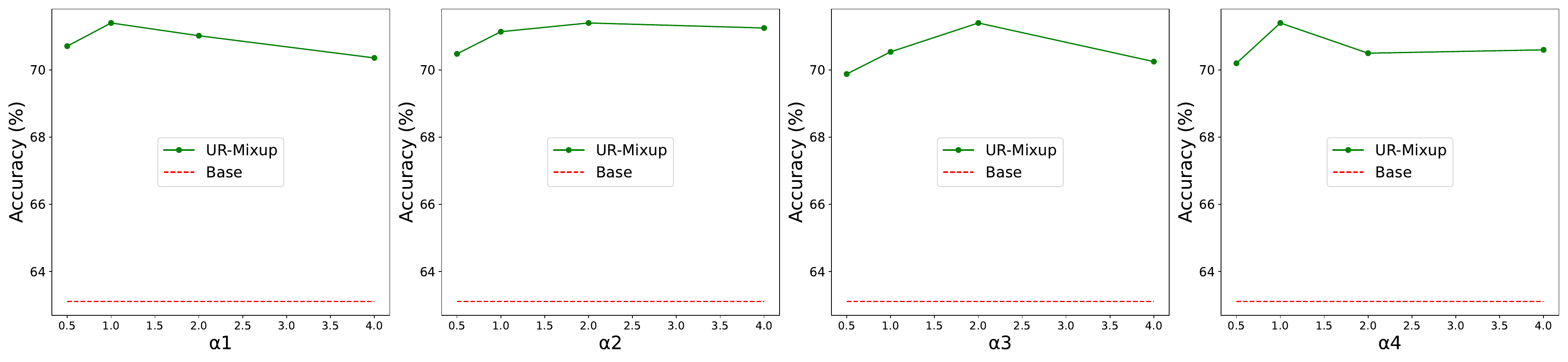}
    \caption{Parameter sensitivity analysis on HAC with video and audio data}
    \label{fig:hyper}
\end{figure*}

\begin{figure*}[h]
    \centering
    \includegraphics[width=1\linewidth]{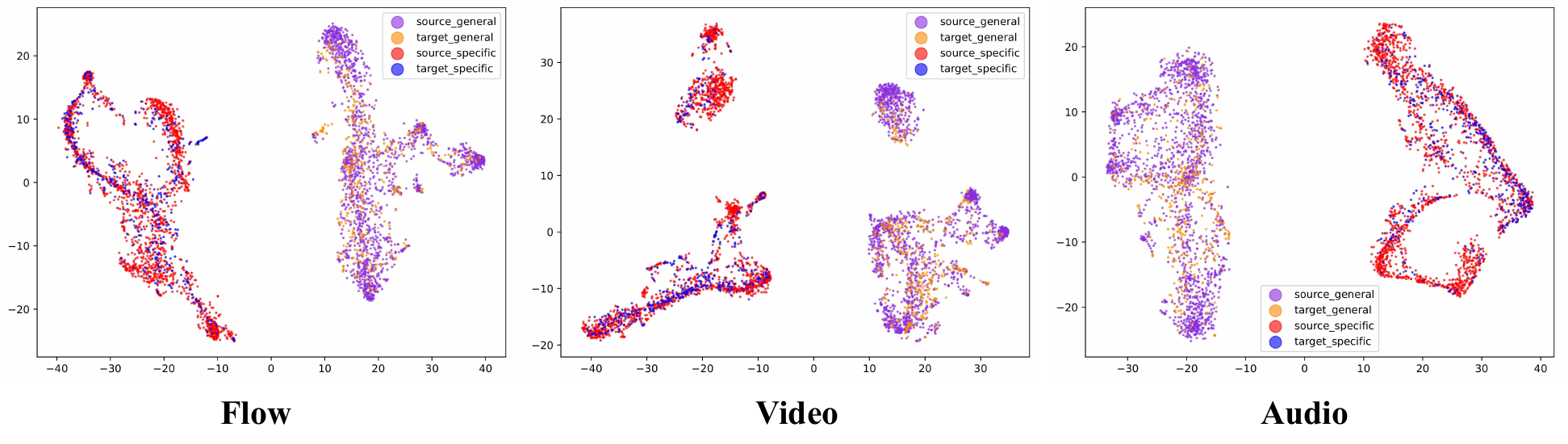}
    \caption{Visualization of the learned embeddings using t-SNE (D2, D3 → D1 in EPIC-Kitchens for multi-modal multi-source DG).}
    \label{fig:decouple_embedding}
\end{figure*}

\noindent
\textbf{More Experiments about recent methods: }
Here, we provide additional results of recent methods with unified representations. Specifically, we report the performance of mDSDI~\cite{bui2021exploiting} and RDM~\cite{nguyen2024domain} under the multimodal multi-source domain generalization setting.

For space efficiency, we report average results across domain splits on EPIC-Kitchens and HAC. As shown in Table~\ref{tab:ur-mdg-compressed}, integrating our unified representation (UR) design with these methods yields consistent and substantial improvements, outperforming the prior SOTA (CMRF) across all modality combinations.

\begin{table}[h]
\centering
\resizebox{\linewidth}{!}{
\begin{tabular}{l|cc|cc|cc|cc}
\hline
\textbf{Method} & \multicolumn{2}{c|}{\textbf{Video+Audio}} & \multicolumn{2}{c|}{\textbf{Video+Flow}} & \multicolumn{2}{c|}{\textbf{Audio+Flow}} & \multicolumn{2}{c}{\textbf{Video+Audio+Flow}} \\
& EPIC & HAC & EPIC & HAC & EPIC & HAC & EPIC & HAC \\
\hline
CMRF & 63.91 & 71.91 & 64.89 & 72.64 & 66.12 & 58.49 & 70.12 & 72.44 \\
mDSDI    & 61.73 & 66.96 & 62.31 & 69.31 & 58.61 & 55.95 & 64.19 & 68.40 \\
UR-mDSDI & 65.61 & 73.65 & 67.25 & 73.76 & 65.99 & \textbf{60.39} & 71.25 & 74.42 \\
RDM      & 62.04 & 67.58 & 62.64 & 69.87 & 58.67 & 56.37 & 63.93 & 68.61 \\
UR-RDM   & \textbf{66.18} & \textbf{74.17} & \textbf{68.07} & \textbf{74.29} & \textbf{66.48} & 59.62 & \textbf{71.88} & \textbf{74.79} \\
\hline
\end{tabular}
}
\caption{Results of mDSDI and RDM with/without UR on EPIC-Kitchens and HAC.}
\label{tab:ur-mdg-compressed}
\vspace{-3mm}
\end{table}

\section{Parameter Sensitivity Analysis}
As shown in Figure~\ref{fig:hyper}, we conduct a comprehensive analysis of four loss hyperparameters in UR-Mixup by varying one parameter at a time while keeping the others fixed. Notably, our method exhibits minimal fluctuations across all parameter settings, indicating a lower sensitivity to hyperparameter selection.

\section{Loss Function Ablation Study}
As shown in Table~5 of the main paper, we ablate $L_{\text{scl}}$ and $L_{\text{club}}$, where rows 1, 3, and 7 correspond to using $L_{\text{cls}}$ only, $L_{\text{cls}} + L_{\text{scl}}$, and the full objective, respectively.  
We do not ablate $L_{\text{cls}}$ since it is essential for classification.  
$L_{\text{rec}}$ is meaningful only when $L_{\text{club}}$ is used, as it ensures semantic completeness after disentanglement.

We further include results using $L_{\text{cls}} + L_{\text{club}}$, $L_{\text{cls}} + L_{\text{scl}} + L_{\text{club}}$, and $L_{\text{cls}} + L_{\text{club}} + L_{\text{rec}}$, in addition to the original settings. Each component contributes positively, confirming its utility in improving performance, as detailed in Table~\ref{tab:loss-ablation}.
\begin{table}[h]
\centering
\resizebox{\linewidth}{!}{
\begin{tabular}{cccc|ccc|c}
\hline
$L_{\text{cls}}$ & $L_{\text{scl}}$ & $L_{\text{club}}$ & $L_{\text{rec}}$ & D2,D3 $\rightarrow$ D1 & D1,D3 $\rightarrow$ D2 & D1,D2 $\rightarrow$ D3 & Mean \\
\hline
\checkmark &         &         &         & 54.94 & 62.26 & 61.70 & 59.63 \\
\checkmark & \checkmark &         &         & 56.24 & 65.38 & 65.07 & 62.23 \\
\checkmark &         & \checkmark &         & 54.75 & 62.53 & 62.08 & 59.79 \\
\checkmark & \checkmark & \checkmark &         & 56.44 & 67.41 & 67.25 & 63.70 \\
\checkmark &         & \checkmark & \checkmark & 55.31 & 63.26 & 63.65 & 60.74 \\
\checkmark & \checkmark & \checkmark & \checkmark & 56.99 & 68.85 & 68.46 & 64.77 \\
\hline
\end{tabular}
}
\caption{Extended ablation study on loss components.}
\label{tab:loss-ablation}
\vspace{-3mm}
\end{table}

\section{More Visualization}
As shown in Figure~\ref{fig:decouple_embedding}, we provide additional visualizations of the learned embeddings. It can be observed that the general and specific information of each modality are well-separated and consistently aligned across domains. Furthermore, the embeddings of flow, video, and audio exhibit strong alignment between the source and target domains.






%% file: table/sup_epic_hac_singlesource.tex
\begin{table}[t!]
\centering
\resizebox{1.0\linewidth}{!}{
\begin{threeparttable}
\begin{tabular}{lcccccc}
\toprule
& \multicolumn{3}{c}{\textbf{Modality}} & \multicolumn{1}{c}{\textbf{EPIC-Kitchens }}& \multicolumn{1}{c}{\textbf{HAC}}\\
\cmidrule(lr){2-4} 
\textbf{Method} & Video & Audio & Flow \\


\midrule

Base & $\checkmark$& $\checkmark$&  & 52.34 &54.18 \\
RNA-Net~\cite{planamente2022domain} & $\checkmark$& $\checkmark$& &  51.25&  54.51 \\
SimMMDG~\cite{dong2023simmmdg}  & $\checkmark$& $\checkmark$&  & 54.84 & 58.75 \\
CMRF~\cite{CMRF}  & $\checkmark$& $\checkmark$&  & \textbf{57.64} & \underline{60.87} \\

\rowcolor{gray!30} 
IBN~\cite{pan2018two}  & $\checkmark$& $\checkmark$&  & 51.43 & 53.62 \\
\rowcolor{gray!30} 
JiGen~\cite{carlucci2019domain}  & $\checkmark$& $\checkmark$& &52.69 & 55.15 \\
\rowcolor{gray!30} 
Mixup~\cite{zhang2017mixup}  & $\checkmark$& $\checkmark$&  & 53.28 &56.29 \\

UR-IBN (ours)  & $\checkmark$& $\checkmark$& & 53.89 & 57.43\\
UR-JiGen (ours)  & $\checkmark$& $\checkmark$&  &56.94 & 60.48 \\
UR-Mixup (ours)  & $\checkmark$& $\checkmark$&  & \underline{57.39} & \textbf{61.32} \\
\midrule

Base & $\checkmark$& &$\checkmark$ &53.64  & 56.80  \\
RNA-Net~\cite{planamente2022domain} & $\checkmark$& & $\checkmark$& 53.86& 57.26  \\
SimMMDG~\cite{dong2023simmmdg}  & $\checkmark$& & $\checkmark$& 57.32 & 60.63 \\
CMRF~\cite{CMRF}  & $\checkmark$& & $\checkmark$& 59.26 &\underline{62.45} \\
\rowcolor{gray!30} 
IBN~\cite{pan2018two}  & $\checkmark$& & $\checkmark$& 54.06 &56.62 \\
\rowcolor{gray!30} 
JiGen~\cite{carlucci2019domain}  & $\checkmark$& & $\checkmark$& 55.64 & 58.23\\
\rowcolor{gray!30} 
Mixup~\cite{zhang2017mixup}  & $\checkmark$& & $\checkmark$& 55.97 & 59.24 \\

UR-IBN (ours)  & $\checkmark$& & $\checkmark$& 55.97 & 58.35\\
UR-JiGen (ours)  & $\checkmark$& & $\checkmark$& \underline{60.21} & 62.44 \\
UR-Mixup (ours)  & $\checkmark$& & $\checkmark$& \textbf{60.46} & \textbf{62.93} \\

\midrule
Base & & $\checkmark$&$\checkmark$ & 48.68 & 44.26\\
RNA-Net~\cite{planamente2022domain} & & $\checkmark$&$\checkmark$ & 49.69 & 42.72  \\
SimMMDG~\cite{dong2023simmmdg}  & & $\checkmark$&$\checkmark$ & 53.27 & 47.28 \\
CMRF~\cite{CMRF}  & & $\checkmark$&$\checkmark$ & \underline{56.46} & 49.96\\
\rowcolor{gray!30} 
IBN~\cite{pan2018two}  & & $\checkmark$&$\checkmark$ & 49.35 & 44.86 \\
\rowcolor{gray!30} 
JiGen~\cite{carlucci2019domain}  & & $\checkmark$&$\checkmark$ & 51.87 & 46.48 \\
\rowcolor{gray!30} 
Mixup~\cite{zhang2017mixup}  & & $\checkmark$&$\checkmark$ & 52.33 & 46.72 \\

UR-IBN (ours)  & & $\checkmark$&$\checkmark$ & 52.23 & 47.51 \\
UR-JiGen (ours)  & & $\checkmark$&$\checkmark$ & \textbf{56.70} & \underline{50.74} \\
UR-Mixup (ours)  & & $\checkmark$&$\checkmark$ & 56.22 & \textbf{51.29} \\

\bottomrule
\end{tabular}

\end{threeparttable}
}
\caption{Multi-modal \textbf{single-source} DG  with different modalities on EPIC-Kitchens and HAC dataset.}
\vspace{-5mm}
\label{tab:sup_epic_hac_singlesource}
\end{table}

%% file: table/epic_compete.tex
\begin{table*}[]
    \centering
    \resizebox{1\linewidth}{!}{
    \begin{tabular}{lccccccccc}
\toprule
 \textbf{Method} & Video & Audio & Video-Audio  & Video & Flow & Video-Flow & Audio & Flow & Audio-Flow\\
\cmidrule(lr){1-1} \cmidrule(lr){2-4} \cmidrule(lr){5-7}  \cmidrule(lr){8-10}
Base (M1) &  58.73 & - & - & 58.73 & - & -& 40.04 & - & -\\
Base (M2)&  - & 40.04 & - & - & 58.30 & -& - & 58.30 & -\\
Base (MM) &  56.65& 38.62& 59.63& 55.28 & 55.78 & 60.89&39.42 & 54.86 & 53.14\\
\midrule
JiGen (M1) & 61.60 & - & - & 61.60 & - & -& 42.72 & - & -\\
JiGen (M2) &  - & 42.72 & - & - & 60.77 & -& - & \underline{60.77} & -\\
JiGen (MM) & 58.98 & 40.67 & 61.08 & 57.14 & 56.64 & 61.79& 40.26 & 56.38 & 58.93\\
\midrule
Mixup (M1) &  61.92& - & - &\underline{61.92}& - & -& \underline{43.74} & - & -\\
Mixup (M2) &  - & \underline{43.74}& - & -& 60.89 & -& - & \textbf{60.89} & -\\
Mixup (MM) &58.52 & 39.31 & 61.18 & 57.86 & 57.24 & 62.08& 40.38 & 57.08 & 58.32\\
\midrule
UR-JiGen (ours) &  \underline{62.02} & 43.41 & \underline{63.63} & 61.26 & \underline{61.08} & \underline{64.15}& 43.52 & \textbf{60.89} & \textbf{63.32}\\
UR-Mixup (ours) &  \textbf{62.45} & \textbf{43.79} & \textbf{64.77} & \textbf{62.34} & \textbf{61.51} & \textbf{66.42} &\textbf{44.24} & 60.24 & \underline{62.63} \\
\bottomrule
    \end{tabular}}
    \caption{The average results of uni-modal performance comparison under multi-modal multi-source DG on EPIC-Kitchens with 3 different modality combinations. M1, M2, and MM denote training settings where the data correspond to the first and second single-modal cases, and the multi-modal case, respectively, following the column header order.
}
    \label{tab:epic_compete}
\end{table*}